\documentclass{article}
\usepackage{spconf,amsmath,graphicx}
\usepackage{amsmath}
\usepackage{amssymb}
\usepackage{bbm}
\usepackage{algorithm}
\usepackage{algpseudocode}
\usepackage{multirow}
\usepackage{subcaption}

\title{Fixed Inter-Neuron Covariability Induces Adversarial Robustness}
\name{Muhammad A. Shah, Bhiksha Raj}
\address{Language Technologies Institute\\
Carnegie Mellon University\\
Pittsburgh, PA}
\begin{document}
\ninept
\maketitle
\begin{abstract}
The vulnerability to adversarial perturbations is a major flaw of Deep Neural Networks (DNNs) that raises question about their reliability when in real-world scenarios. On the other hand, human perception, which DNNs are supposed to emulate, is highly robust to such perturbations, indicating that there may be certain features of the human perception that make it robust but are not represented in the current class of DNNs. One such feature is that the activity of biological neurons is correlated and the structure of this correlation tends to be rather rigid over long spans of times, even if it hampers performance and learning. We hypothesize that integrating such constraints on the activations of a DNN would improve its adversarial robustness, and, to test this hypothesis, we have developed the Self-Consistent Activation (SCA) layer, which comprises of neurons whose activations are consistent with each other, as they conform to a fixed, but learned, covariability pattern. When evaluated on image and sound recognition tasks, the models with a SCA layer achieved high accuracy, and exhibited significantly greater robustness than multi-layer perceptron models to state-of-the-art Auto-PGD adversarial attacks \textit{without being trained on adversarially perturbed data}.
\end{abstract}
\newcommand{\Reps}[1]{\mathbf{R}_{#1}}
\newcommand{\xb}{\mathbf{x}}
\newcommand{\ab}{\mathbf{a}_\xb}
\newcommand{\ub}{\mathbf{u}}
\newcommand{\Wb}{\mathbf{W}}
\newcommand{\gC}{g_{\mathcal{C}}}
\newcommand{\hX}{h_{\mathcal{X}}}
\newcommand{\dimx}{d_{\xb}}
\newcommand{\dima}{d_{\mathbf{a}}}
\newcommand{\Rdx}{\mathbbm{R}^{\dimx}}
\newcommand{\Rda}{\mathbbm{R}^{\dima}}

\newcommand{\Wbf}{\mathbf{W}_f}
\newcommand{\Wbg}{\mathbf{W}_g}
\newcommand{\Wbh}{\mathbf{W}_h}
\newcommand{\bbf}{\mathbf{b}_f}
\newcommand{\bbg}{\mathbf{b}_g}
\newcommand{\bbh}{\mathbf{b}_h}
\newcommand{\actfn}{\phi}
\newcommand{\norm}[1]{\|#1\|}
\newcommand{\sqnorm}[1]{\norm{#1}_2^2}
\newcommand{\fnorm}[1]{\norm{#1}_F}

\newcommand{\fwdTerm}{\actfn(\Wbf\xb + \bbf)}
\newcommand{\latTerm}{\actfn(\Wbg\ab + \bbg)}
\newcommand{\bkTerm}{\Wbh\xb + \bbh}
\newcommand{\linf}{\ell_{\infty}}

\section{Introduction}
\label{sec:intro}
While Deep Neural Networks (DNNs) have advanced the state of the art in many fields of Artificial Intelligence (AI), their vulnerability to adversarial perturbation -- subtle distortions that are semantically irrelevant to humans but can change the response of DNNs when added to their inputs \cite{goodfellow13, ilyas2019adversarial} -- raises concerns about their reliability when deployed in high-stakes real world applications such as self-driving cars, and biometric authentication systems. These concerns will need to be resolved for the wide-spread integration of AI into human society to be possible.

Over the years, the development of methods to make DNNs more robust to adversarial perturbations has become an active area of research. A large fraction of the proposed methods rely heavily on machine learning and statistical analysis techniques \cite{madry2017towards,cohen2019certified,salman2020denoised,anonymous2023less,shah2021towards,olivier2021high}, with the most popular and successful of them being adversarial training in which involves training on adversarially perturbed data \cite{madry2017towards}. Despite their effectiveness, these methods have two major shortcommings: they trade-off accuracy on natural data for adversarial robustness, and their effectiveness deteriorates if presented with perturbations other than the precise type of adversarial perturbation that they are designed to defend against \cite{joos2022adversarial,sharma2017attacking,schott2018towards}, for instance a technique designed to defend against low-magnitude perturbations applied to the whole input may not be effective against high-magnitude perturbations applied to a very localized region of the input \cite{joos2022adversarial}.

In contrast, human perception is highly accurate and naturally robust to a wide variety of such perturbations \textit{without ever being exposed to them}, and therefore it may be beneficial, as has been the case throughout the history of neural networks, to seek inspiration from biological perceptual systems. Following this reasoning, several biologically-inspired methods have been proposed for defending DNNs against adversarial perturbations. These methods generally computationalize, and integrate into DNNs, some mechanism of biological perception that is as yet unrepresented in modern DNNs, but is hypothesized to contribute to its robustness \cite{paiton2020selectivity,dapello2020simulating,jonnalagadda2022foveater,vuyyuru2020biologically}. The improvement in robustness that these methods yield is relatively modest, compared to the non-biologically inspired methods mentioned earlier, however, it usually does not come at the cost of accuracy and tends to generalize better to other types of perturbations. 

Building on this body of work, in this paper we investigate the role of inflexible inter-neuron covariability structures in making perception more robust. It has been observed that in the brain the spiking activity of individual neurons tends to be correlated \cite{hennig2021learning, sadtler2014neural} and, the structure of this correlation tends to be inflexible over long periods of time even if it limits performance and learning \cite{golub2018learning}. We hypothesize that integrating such a mechanism into DNNs may improve their robustness because it will restrict the space of adversarial perturbations to only those that give rise to activations that respect a fixed covariability structure. In other words, forcing the intermediate activations to conform to a fixed covariability structure prevents the adversarial perturbation from causing arbitrary changes to the intermediate activations and thus misclassifications.



Integrating this behaviour into DNNs is not straightforward, because, 
unlike biological neurons, neurons in a DNN 
output a deterministic real number, which raises the question of how can we simulate correlated spiking in a system that is neither stochastic nor produces spiking activity? A solution presents itself if we consider the outputs of the artificial neuron as the frequency of an underlying spike train. If the spiking activity of a group of neurons is correlated, then the frequency of their spikes may also be correlated. Therefore, we use spiking frequency as a proxy for the spikes trains and, since the spiking frequency is represented in a DNN by the outputs of the neurons, we impose a fixed covariability structure on the latter.


To simulate a fixed inter-neuron covariability pattern, we have developed the Self-Consistent Activation (SCA) layer, which comprises of neurons whose activations are consistent with each other as they conform to a fixed covariability pattern. The SCA layer first computes the feed forward activations for the neurons based on the input, and then iteratively optimizes these activations to make them conform to a fixed, but learned, covariability pattern.

We evaluated the effectiveness of the SCA layers on image and sound classification tasks. For image classification we used MNIST \cite{lecun2010mnist} and Fashion-MNIST (FMNIST) \cite{xiao2017fashion}, and for sound classification we used SpeechCommands \cite{warden2018speech}. 
Compared to a  Multi-Layer Perceptron (MLP), we find that the inter-neuron correlations in models with a SCA layer are more invariant to adversarial peturbations, thus indicating that the SCA layer does indeed make the inter-neuron covariance structure more inflexible as we intended. 
Furthermore, models with the SCA layer achieved similar if not better accuracy on the unperturbed data, and \textit{significantly higher} accuracy on adversarially perturbed data generated by Auto-PGD \cite{croce2020reliable} -- a state-of-the-art  white box adversarial attack. We evaluated the accuracy of the models under adversarial perturbations of several sizes and found that on average the model with the SCA layer achieved an absolute improvement of 4\%, 5\% and 6\%, and a relative improvement of 117\%, 155\%
and 45\%, compared to the MLP model on FMNIST, SpeechCommands and MNIST, respectively. Similar trends were observed when the models were trained on adversarially perturbed data, with the SCA model achieving significantly higher accuracy on the clean and adversarially perturbed data on all datasets except MNIST. 

\section{Related Work}
\textbf{Non-Biologically Inspired Methods for Robustness:}
Most of the proposed methods for making DNNs robust to adversarial perturbations are not inspired from biology but rather rely on machine learning and statistical analysis techniques. Perhaps the most successful of these methods is adversarial training \cite{madry2017towards, bai2021recent}, which adversarially perturbs each training minibatch by using Projected Gradient Descent to modify the data in the minibatch such that the loss is maximized.
Over the years several improvements to the basic adversarial training algorithm have been proposed, with each modifying different parts of the model training pipeline, such as data augmentation \cite{rebuffi2021data}, and regularization \cite{zhang2019theoretically}. Another popular group of methods concentrate on creating models that are provably robust against adversarial perturbations and are accompanied by formal guarantees of the form: with probability at least $1-\delta$, where $\delta$ is small, the model's output will not change if a perturbation having norm at most $\epsilon$ is added to a given input \cite{cohen2019certified, fischer2020certified, kumar2021center, li2019certified}. 
These methods have two major shortcommings: they trade-off accuracy on natural data for adversarial robustness, and their effectiveness deteriorates if presented with perturbations other than the precise type of adversarial perturbation that they are designed to defend against \cite{joos2022adversarial,sharma2017attacking,schott2018towards}. 

\textbf{Biologically Inspired Methods for Robustness:}
These methods generally involve developing computational analogues of biological process that are absent from common DNNs. Examples of such processes are predictive coding \cite{paiton2020selectivity,bai2021recent}, associative memory \cite{krotov2018dense} biologically constrained visual filters, nonlinearities and stochasticity \cite{dapello2020simulating}, foveation \cite{jonnalagadda2022foveater, luo2015foveation, gant2021evaluating,anonymous2023less}, and non-uniform retinal sampling and cortical fixations \cite{vuyyuru2020biologically}. Perhaps most closely related to our work are the methods based on predictive coding theory, which posits that the brain is a hierarchical Bayesian network in which the output of each layer is the maximum aposteriori estimate of the activations given the output of the previous layer and the next layer \cite{rao1999predictive}. Integrating predictive coding layers within a model has been shown to improve its adversarial robustness \cite{paiton2020selectivity,choksi2021predify}. This approach is similar to ours in so far as 
it maximizes some notion of consistency between activations. Under predictive coding the activations become consistent when the input can be reconstructed perfectly from them, while in our work they become consistent when their covariability structure matches the one learned during training.

\section{The Covariability of DNN Activations}
\label{sec:covar_analysis}
\begin{figure}
     \centering
     \begin{subfigure}[b]{.21\textwidth}
         \centering
         \includegraphics[width=\textwidth]{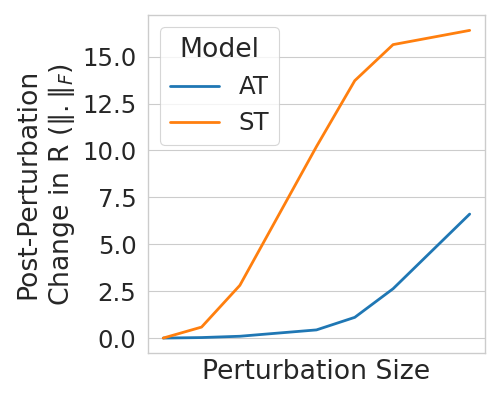}
         \caption{}
         \label{fig:mlp64_corr_norm}
     \end{subfigure}
     \begin{subfigure}[b]{.21\textwidth}
         \centering
         \includegraphics[width=\textwidth]{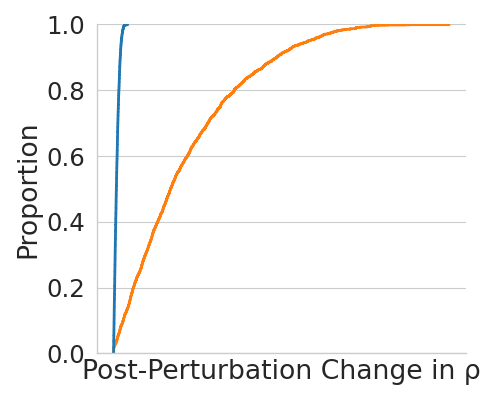}
         \caption{}
         \label{fig:mlp64_corr_cdf}
     \end{subfigure}
        \caption{
        (a) The Frobenius norm of the change in the correlation matrix of the activations of neurons in the penultimate layer of a MLP trained on clean (ST) and adversarially perturbed (AT) FMNIST images when adversarial perturbations of different sizes are added to the input; (b) CDF of the change in correlation between neuron pairs when adversarial perturbation of $\linf$ norm 0.1 is added.
        }
        \label{fig:mlp64_corr}
        \vspace{-10px}
\end{figure}

We hypothesize that the inflexible covariability structure of neuronal activations that is observed in the animal brain contributes to the robustness of biological vision. As a preliminary step in our investigation of this hypothesis we determine if there is a relationship between the inflexibility of the correlation matrix of a DNN's activations, which we consider a proxy of its covariability structure, and its robustness to adversarial perturbations. To this end, we analyse the correlation structure between the neural activations of a DNN in response to data perturbed with perturbations of different sizes. To this end, we train two 5-layer MLP models on FMNIST, one on clean data and the other via adversarial training, and compute the correlation between the activations of the penultimate layer in response to clean and adversarially perturbed images. We use $\Reps{\epsilon}$ to refer to the correlation matrix produced by data perturbed by perturbations of $\linf$ norm $\epsilon$. We quantify the overall change in the correlation structure as $\fnorm{\Reps{0}-\Reps{\epsilon}}$, where $\fnorm{\cdot}$ is the Frobenius norm and plot this quantity for several values of $\epsilon$ in Figure \ref{fig:mlp64_corr_norm}. We can see that the correlation structure of the adversarially trained MLP is much more invariant to adversarial perturbations compared to the correlation structure of the MLP trained on clean data. It is only after the size of the perturbation become very large does the correlation structure of the adversarially trained model begin to change significantly. To verify that the change in the norm is not caused due to a small number of neurons, we compute the absolute change in the correlation of each neuron pair due to the addition of adversarial perturbations of size 0.1, and plot the cumulative frequency curve shown in Figure \ref{fig:mlp64_corr_cdf}. We see that the curve for the adversarially trained model is significantly shifted to the left of the curve for the model trained on clean data indicating that the correlation between most, if not all, the pairs of neurons has not changed significantly. 

From these observations we can infer that  the invariance of the inter-neuron covariability structure across different perturbations of the input is related to the adversarial robustness of the model. If this relationship is causal then constraining the inter-neuron covariability structure should induce adversarial robustness. To determine if this is indeed the case we design a neural network layer that explicitly optimizes its activations to make them conform to a fixed covariability structure. We then include this layer in a DNN model and evaluate its robustness against state-of-the-art adversarial attacks.


\section{Self-Consistent Activation Layer}
\label{sec:sca-layer}
\newcommand{\C}{\mathcal{C}}
\newcommand{\A}{\mathcal{A}}
\newcommand{\X}{\mathcal{X}}

We have developed the Self-Consistent (SCA) Activation Layer to simulate an inflexible inter-neuron covariability structure. At a high-level, the SCA layer computes its output as ${\text{SCA}(\xb) = \gC(\ab)}$
where $\xb\in\Rdx$ is the input, $\ab=f(\xb)\in\Rda$ is the feed-forward activation vector, and $\gC(\ab)$ is the projection of $\ab$ onto $\C$, the subspace comprised of the vectors that respect the learned covariance structure.
If we consider covariability to be a linear relationship, like covariance, then $\gC$ would simply be a linear projection.
However, to allow for more complex inter-neuron interaction, in this paper we have decided to adopt the following non-linear form for $\gC$:
\begin{align}
    \arg\min_{\ab}&\sqnorm{\ab - \latTerm} + \lambda\sqnorm{\xb - \Wbh\ab - \bbh},
    \label{eq:sca-loss}
\end{align}
where $\phi=ReLU$, $\Wbg\in\mathbbm{R}^{\dima\times\dima}$, $\Wbh\in\mathbbm{R}^{\dimx\times\dima}$, $\bbg\in\Rda$ and $\bbh\in\Rdx$. The first term represents the distance between the activation and its projection onto $\C$, while the second term represents the information about $\xb$ that is not carried by $\ab$. The latter is added as a regularizer to prevent degenerate solutions, like $\ab=0$, in which $\ab$ carries no information about $\xb$, and $\lambda$ is a scalar that controls the strength of the regularization. We set the diagonal of $\Wbg$ to zero to prevent it from becoming the identity matrix and we perform the minimization using batch gradient descent. The exact sequence of operations performed by the SCA layer is shown in Algorithm \ref{alg:sca-layer}.

\begin{algorithm}
    \caption{SCA Layer}\label{alg:sca-layer}
    \begin{algorithmic}[1]
        \State $\ub \gets f(\xb)$
        \For{$t: 1\rightarrow T$}
            \State $\ab \gets \actfn(\ub)$
            \State $J\gets \sqnorm{\ab - \latTerm} +\lambda\sqnorm{\xb - \Wbh\ab - \bbh}$
            \State $\ub \gets \ab - \eta\nabla_{\ab}J$
        \EndFor
        \State $\ab \gets \actfn(\ub)$
    \end{algorithmic}
    
\end{algorithm}

\section{Evaluation}
\subsection{Experimental Setup}
\subsubsection{Datasets}
We evaluate the performance of SCA layers on image and audio classification tasks. For image classification we use the MNIST \cite{lecun2010mnist} and Fashion MNIST (FMNIST) \cite{xiao2017fashion} datasets, which contain 60,000 $28\times28$ black-and-white images of handwritten digits and 10 types of clothes, respectively. From both MNIST and FMNIST, we used 45,000 images for training, 5000 for evaluation and 10,000 for testing. For the audio classification task we use a subset of SpeechCommands dataset \cite{warden2018speech}, which contains about 40,000 1 second, 16KHz recordings of humans vocalizing digits 0 to 9. We use 31,000 recordings for training, 4,000 recordings for validation and 4,000 recordings for testing.

\subsubsection{Data Preprocessing}
For the image datasets, we flatten the image into a vector, which is then normalized 
by subtracting 0.5 and then dividing by 0.5. The audio data is pre-processed by first downsampling to 8KHz. Then 128 Mel-Frequency Cepstral Coefficients (MFCCs) are computed from a log mel spectrogram having 512 FFT points computed over a 64 ms sliding window with a stride of 32ms. By retaining only the first 16 MFCCs we obtain a $16\times 251$ matrix for each 1s audio recording. The matrix is then flattened, and normalized by subtracting -0.96 and then dividing by 9.2 (the mean and standard deviation computed over the validation set).


It is important to note here that the adversarial perturbations are computed on and applied to the original input, prior to pre-processing. All the above mentioned pre-processing steps are implemented using PyTorch and torchaudio, and are fully differentiable, therefore the gradients can be propagated back to the original input and used to compute adversarial perturbations. 

\subsubsection{Models}
\begin{figure}
    \centering
    \includegraphics[width=.4\textwidth]{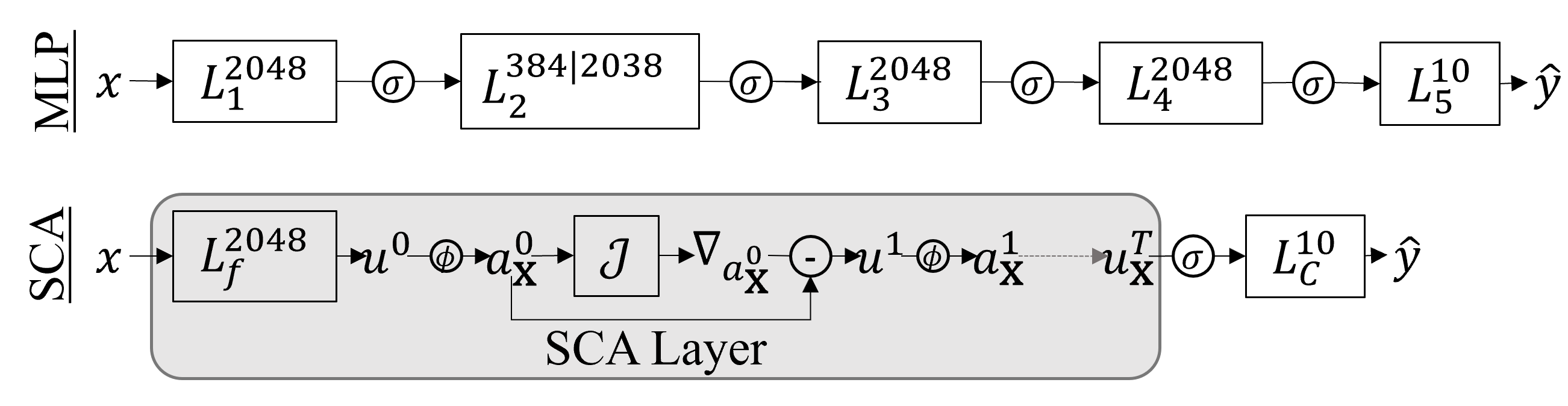}
    \caption{Schematics of the MLP and SCA models. $L_.$s are affine projections with the superscripts representing the output dimension. The output dimension of $L_2$ is set to 384 in MLPs trained on MNIST and FMNIST, and to 2048 in MLPs trained on SpeechCommands. $\phi=ReLU$, $\sigma=\text{dropout}\circ \phi$, $\mathcal{J}$ is the loss from eq. (\ref{eq:sca-loss}).}
    \label{fig:model_arch}
\end{figure}

We compare the performance of models containing SCA layers (SCA model) to MLP models having comparable architectures and number of parameters. The schematics of these models are shown in Figure \ref{fig:model_arch}. The SCA model performs $T=16$ optimization steps. The probability of dropout is set to 0.2 for all models.

The models are optimized using the Adam optimizer using a learning rate of 0.001 and a batch size of 256 for up to 100 epochs. The learning rate is halved if the loss on the validation set does not decrease for 5 epochs, and if it does not decrease for 20 epochs the training is stopped early. All the results presented below are averaged over 5 trials with different random seeds. 

\subsection{Results}
\subsubsection{Analysis of Activation Covariability Structure}
\begin{figure}
     \centering
     \begin{subfigure}[b]{.15\textwidth}
         \centering
         \includegraphics[width=\textwidth]{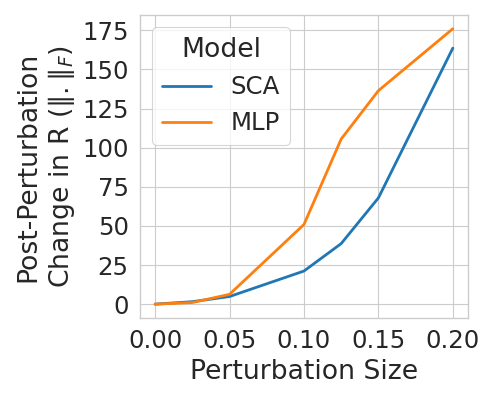}
         \caption{MNIST}
         \label{fig:mlp2048_mnist_corr_norm}
     \end{subfigure}
     \begin{subfigure}[b]{.15\textwidth}
         \centering
         \includegraphics[width=\textwidth]{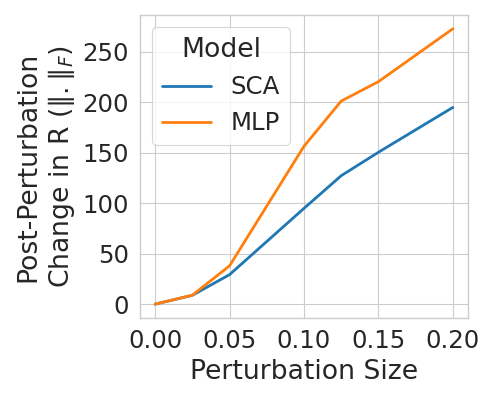}
         \caption{FMNIST}
         \label{fig:mlp2048_fmnist_corr_norm}
     \end{subfigure}
     \begin{subfigure}[b]{.15\textwidth}
         \centering
         \includegraphics[width=\textwidth]{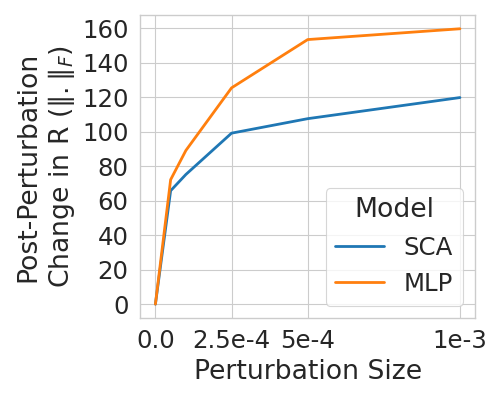}
         \caption{SpeechCommands}
         \label{fig:mlp2048_sc_corr_norm}
     \end{subfigure}
        \caption{The Frobenius norm of the change in the correlation matrix of the penultimate layer activations from the SCA model and MLP due to the addition of adversarial perturbations of different $\linf$ norms.}
        \label{fig:mlp2048_corr}
\end{figure}

To verify that SCA layers increase the invariance of the inter-neuron correlation structure we analyse the correlation between the activations of the penultimate layer using the method introduced in \ref{sec:covar_analysis}. Specifically, we compute the correlation matrix $\Reps{\epsilon}$ from the activations of the penultimate layer of the SCA model and MLP in response to 1000 data samples, perturbed by adversarial perturbations of $\linf$ norm $\epsilon$. We compute $\Reps{\epsilon}$ for each dataset using several values of $\epsilon$. For each dataset and $\epsilon$ we then compute $\fnorm{\Reps{0}-\Reps{\epsilon}}$ to represent the overall change in the correlation structure due to the addition of adversarial perturbation of size $\epsilon$. Figure \ref{fig:mlp2048_corr} shows this quantity for the SCA model and MLP on each dataset. We see that in every case the correlation structure of the SCA model changes more slowly than the MLP, and thus is more invariant to adversarial perturbation. This result shows that the SCA layer indeed produces the intended effect of constraining the covariability structure of neural activations.

\subsubsection{Robustness of Models Trained on Clean Data}
We evaluate the adversarial robustness of the SCA models by training them on \textit{clean} data and computing their classification accuracy on adversarially perturbed data. The adversarial perturbations are computed by running AutoAttack \cite{croce2020reliable} with different bounds on the maximum $\linf$ norm of the perturbation. 

The accuracy achieved by the SCA models and the baseline MLPs under adversarial attack is shown in Table \ref{tab:clean-train}. We see that on all the three datasets the SCA model is \textit{significantly} more robust than the MLP to adversarial perturbations, achieving higher accuracy on perturbations of all sizes. Averaged across all perturbation sizes, the SCA model achieves an absolute improvement in accuracy of 4\%, 5\% and 6\%, and a relative improvement of 117\%, 155\% and 45\%, compared to the MLP model on FMNIST, SpeechCommands and MNIST, respectively. The above results clearly show that SCA layers are much more robust to adversarial attacks than layers of perceptrons, and that their robustness is not limited to a particular type of data but generalizes across data complexity and modality.

\newcommand{\sig}{$^*$}
\begin{table}[]
    \centering
    \begin{tabular}{c c c c c c c}
    \hline
    Model & \multicolumn{6}{c}{Perturbation Sizes ($\linf$)}\\
    & 0.0 & 0.05 & 0.1 & 0.125 & 0.15 & 0.2\\\hline
    & \multicolumn{6}{c}{MNIST}\\
    SCA & 97.9 & 88.1 & \textbf{54.8} & \textbf{32.3} & \textbf{15.7} & \textbf{1.7}\\
    MLP & \textbf{98.4} & \textbf{90.0} & 52.6 & 28.2 & 12.8 & 1.4\\\hline
    & \multicolumn{6}{c}{FMNIST}\\
    SCA & \textbf{89.4}\sig & \textbf{46.9}\sig & \textbf{12.7}\sig & \textbf{6.1}\sig & \textbf{2.8}\sig & \textbf{0.1}\\
    MLP & 88.7 & 39.2 & 7.7 & 3.1 & 1.0 & 0.0\\\hline
    & \multicolumn{6}{c}{Perturbation Sizes ($\linf$)}\\
    & 0.0 & 5e-5 & 1e-4 & 2.5e-4 & 5e-4 & 1e-3\\\hline
    & \multicolumn{6}{c}{SpeechCommands}\\
    SCA & \textbf{90.1}\sig & \textbf{45.1}\sig & \textbf{32.2}\sig & \textbf{10.3}\sig & \textbf{2.4}\sig & \textbf{0.5}\sig\\
    MLP & 88.2 & 38.4 & 20.7 & 3.7 & 0.8 & 0.1\\\hline
    \multicolumn{6}{c}{} & \sig$p<0.05$
    \end{tabular}
    \caption{The accuracy (\%) achieved by MLP and the SCA models}
    \label{tab:clean-train}
    \vspace{-10px}
\end{table}

\subsubsection{Robustness of Adversarially Trained Models}
Given the SCA model is naturally more robust to adversarial perturbations, we hypothesize that an adversarially trained SCA model will also be more robust than an adversarially trained MLP. To verify this hypothesis, we adversarially train MLP and SCA models using the method from \cite{madry2017towards}: for each mini-batch we computer adversarial pertuabtions having $\linf$-norm at most $\epsilon$ by performing 7 iterations of PGD with the step size $\epsilon/4$, where $\epsilon$ is set to 0.3 for MNIST and FMNIST, and 2e-4 for SpeechCommands. 

Table \ref{tab:adv-train} shows the accuracy of the adversarially trained models under Auto-PGD attack \cite{croce2020reliable}. The SCA model achieves greater accuracy than the MLP model for all perturbation sizes on SpeechCommands, and for perturbation size greater than 0.3 on FMNIST. Averaged across all perturbation sizes, the SCA model achieves an absolute improvement in accuracy of 1.7\% and 2.4\%, and a relative improvement of 49\% and 12.6\%, compared to the MLP model on FMNIST and SpeechCommands, respectively.  On the other hand, we find that the adversarially trained MLP performs better than the SCA model under large perturbations on MNIST. 
The fact that the SCA model is significantly more robust than the MLP on more complex datasets, particularly SpeechCommands, but not on the simpler MNIST, may indicate that, under adversarial training, the utility of SCA layers is realized only in more complex tasks.


\begin{table}[]
    \centering
    \begin{tabular}{c c c c c c c}
    \hline
    Model & \multicolumn{6}{c}{Perturbation Sizes ($\linf$)}\\
    & 0.0 & 0.2 & \textbf{0.3} & 0.325 & 0.35 & 0.375\\\hline
    & \multicolumn{6}{c}{MNIST}\\
    SCA & \textbf{96.9}\sig & \textbf{88.0}\sig & 70.4 & 56.3 & 29.9 & 7.9\\
    MLP & 94.7 & 85.8 & \textbf{74.9}\sig & \textbf{64.6}\sig & \textbf{34.4} & \textbf{11.0}\sig\\\hline
    & \multicolumn{6}{c}{FMNIST}\\
    SCA & \textbf{70.4}\sig & 51.5 & 35.1 & \textbf{22.4}\sig & \textbf{7.0}\sig & \textbf{2.3}\sig\\
    MLP & 67.8 & \textbf{53.5}\sig & \textbf{42.4}\sig & 10.6 & 3.1 & 1.3\\\hline
    & \multicolumn{6}{c}{Perturbation Sizes ($\linf$)}\\
    & 0.0 & 1e-4 & \textbf{2.5e-4} & 5e-4 & 1e-3 & 2e-3\\\hline
    & \multicolumn{6}{c}{SpeechCommands}\\
    SCA & \textbf{67.3} & \textbf{61.2}\sig & \textbf{56.4}\sig & \textbf{46.9}\sig & \textbf{30.9}\sig & \textbf{14.2}\sig\\
    MLP & 66.3 & 59.6 & 54.6 & 43.6 & 26.8 & 11.4\\\hline
    \multicolumn{6}{c}{} & \sig$p<0.05$
    \end{tabular}
    \caption{The accuracy achieved by adversarially trained MLP and SCA models under adversarial perturbations of various $\linf$ norms. The norm of the perturbations used during training is 0.3 for MNIST and FMNIST models, and 2.4e-4 for the SpeechCommands models.}
    \label{tab:adv-train}
\end{table}

\subsubsection{Impact of Number of Steps}
To determine the impact of changing the number of steps, $T$, in Algorithm \ref{alg:sca-layer}, we train models with different values of $T$ using only clean data from FMNIST. The accuracy of these models on clean and perturbed data is shown in Table \ref{tab:acc_v_steps}. We observe that increasing $T$ does not impact clean accuracy but robustness significantly improves as $T$ is increased beyond 8. This indicates that the more self-consistent the activations become the more robust the model gets.
\begin{table}[]
    \centering
    \begin{tabular}{c c c c c c}
        \hline
        & \multicolumn{5}{c}{Number of Steps}\\
        & 1 & 2 & 4 & 8 & 16\\\hline
        clean & 89.3 & 89.5 & 89.8 & 89.1 & 89.5\\
        perturbed & 35.8 & 35.8 & 35.7 & 40.3 & 49.3\\\hline
    \end{tabular}
    \caption{The accuracy of SCA models with different number of self-consistency optimizing steps ($T$) on clean and adversarially perturbed data. The size of adversarial perturbations is set to 0.05.}
    \label{tab:acc_v_steps}
    \vspace{-10px}
\end{table}

\section{Conclusion}
In this paper we investigate the impact of the biological phenomenon of inflexible inter-neuron covariability on robustness to adversarial perturbation. To this end we develop the SCA layer, which as neural network layer that comprises of neurons whose activations conform to a fixed, but learned, covariability pattern. We demonstrate that DNNs with the SCA layer tend to be significantly more robust to adversarial perturbations than conventional MLPs \textit{without ever being trained on adversarially perturbed data}, achieving an average improvement of 102\%  in accuracy relative to the MLP on image and audio data perturbed by state-of-the-art adversarial attack methods. These results indicate that constraining the inter-neuron covariability structure does indeed make DNNs more robust. More generally, these results lend credence to the approach of seeking inspiration from biology for developing better and more reliable AI systems.
\bibliographystyle{IEEEbib}
\bibliography{refs}

\end{document}